\documentclass[11pt]{article}
\usepackage{dialogue2021}

\usepackage{ifxetex}
\ifxetex
    \usepackage{fontspec}
    \setromanfont{Times New Roman}
\else
  \usepackage[T1]{fontenc}
  \usepackage[utf8]{inputenc}
  \usepackage{cmap}
  \usepackage{times}
  \usepackage{latexsym}

\fi

\usepackage[russian,british]{babel}
\usepackage{url}
\usepackage{pgf}

\usepackage{covington} 

\dialogfinalcopy 

\title{Transfer Learning for Improving Results on Russian Sentiment Datasets}

\author{Anton Golubev \\
  Bauman Moscow State \\
  Technical University \\
  Russia \\
  {\tt antongolubev5@yandex.ru} \\\And
  Natalia Loukachevitch \\
  Lomonosov Moscow State \\ 
  University \\
  Russia \\
  {\tt louk\_nat@mail.ru} \\}

\date{}

\begin{document}
\maketitle
\begin{abstract}
 In this study, we test transfer learning approach on Russian sentiment benchmark datasets using additional train sample created with distant supervision technique. We compare several variants of combining additional data with benchmark train samples. The best results were achieved using three-step approach of sequential training on general, thematic and original train samples. For most datasets, the results were improved by more than 3\% to the current state-of-the-art methods. The BERT-NLI model treating sentiment classification problem as a natural language inference task reached the human level of sentiment analysis on one of the datasets.
  
  
  \textbf{DOI:} 10.28995/2075-7182-2021-20-XX-XX
\end{abstract}

\section{Introduction}
Sentiment analysis or opinion mining is an important natural language processing task used to determine sentiment attitude of the text. One of its main business application is product monitoring consisting of studying customer feedback and needs. Nowadays most  state-of-the-art results are obtained using deep learning models, which require training on specialized labeled data.

In recent years transfer learning has earned widespread popularity. This approach includes a pre-training step of learning general representations from a source task and an adaptation step of applying previously gained knowledge to a target task. In other words, deep learning model trained for a task is reused as the starting point for a model on a second task. Since there is a significant amount of text information nowadays, current state-of-the-art results can be possibly improved using transfer learning.

The most known Russian sentiment analysis datasets include ROMIP-2013 and \newline SentiRuEval2015-2016 \cite{chetviorkin2013evaluating,loukachevitch2015entity,loukachevitch2016rubtsova}, which consist of annotated data on banks and telecom operators reviews from Twitter messages and news quotes. Current best results on these datasets were obtained using pre-trained RuBERT \cite{smetanin2021deep,golubev2020loukachevitch} and conversational BERT model \cite{devlin2018bert,deeppavlov} fine-tuned as architectures treating a sentiment classification task as a natural language inference 
(NLI) or question answering (QA) problem \cite{golubev2020loukachevitch}. 

In this study, we introduce a method for automatic generation of annotated sample from a Russian news corpus using distant supervision technique. We compare different variants of combining additional data with original train samples and test the transfer learning approach based on several BERT models. For most datasets, the  results  were  improved  by  more  than  3\%  to  the  current state-of-the-art performance. On SentiRuEval-2015 Telecom Operators Dataset, the BERT-NLI model treating a sentiment classification problem as a natural language inference task, reached human level according to one of the metrics.

The contributions of this paper are presented below:
\begin{itemize}
    \item  we propose a new method  of automatic generation of additional data for sentiment analysis tasks from raw texts using the distant supervision approach and a sentiment lexicon,
    \item we compare several variants of combining the additional dataset with original train samples and show that three-step approach of sequential training on general, thematic and benchmark train samples performs better, 
    \item we renew the best results on five Russian sentiment analysis datasets using pre-trained BERT models combined with transfer learning approach,
    \item we show that BERT-NLI model treating sentiment classification problem as a natural language inference task reaches human level on one of the datasets.
\end{itemize}

This paper is structured as follows. In Section 2, we overview related methods applied to the considered task. Section 3 describes sentiment analysis datasets used in this paper. Section 4 represents a process of automatic generation of an annotated dataset using distant supervision approach. In section 5 and 6 we briefly cover main preprocessing steps and BERT models applied in the current study. Section 7 presents the full track of transfer learning study including comparison of different variants of constructing an additional datasets and several ways of their combining with benchmark train samples. 

\section {Related Work}
Russian sentiment analysis datasets are based on different data sources  \cite{smetanin2021deep}, including reviews \cite{smetanin2019sentiment,chetviorkin2013evaluating}, news stories \cite{chetviorkin2013evaluating}, posts from social networks \cite{rubtsova2015constructing,loukachevitch2015entity,rogers2018rusentiment}. The best results on most available datasets are obtained using transfer learning approaches based on the BERT  model \cite{devlin2018bert}, more specifically on RuBERT \cite{deeppavlov} and Russian variant of BERT \cite{smetanin2021deep,golubev2020loukachevitch,moshkin2020application,baymurzina2019language}. In \cite{golubev2020loukachevitch}, the authors tested several variants of RuBERT and different settings of its applications, and found that the best results on sentiment analysis tasks on several datasets were achieved using Conversational RuBERT trained on Russian social networks posts and comments. Among several architectures, the BERT-NLI model treating the sentiment classification problem as a natural language inference task usually has the highest results. 

For automatic generation of annotated data for sentiment analysis task, researchers use so-called distant supervision approach, which exploits  additional resources:  users' tags or manual lexicons \cite{go2009twitter,rubtsova2015constructing}. For Twitter sentiment analysis, users' positive or negative emoticons or hashtags can be used \cite{sahni2017efficient,rubtsova2015constructing,mohammad2016sentiment}.
Authors of \cite{rusnachenko2019distant} use the RuSentiFrames lexicon for creating a large automatically annotated dataset for recognition of sentiment  relations between mentioned entities.

In contrast to previous work, in the current study we automatically create a dataset for targeted sentiment analysis, which extracts a sentiment attitude towards a specific entity. The use of an automatic dataset together with manually annotated data allows us to improve the state-of-the-art results. 

\begin{table}[h]
\centering
\caption{Benchmark sample sizes and sentiment class distributions (\%).}\label{distributions}
\begin{tabular}{|l|c|c|c|c|c|c|c|c|}
\hline
& \multicolumn{4}{c|}{Train sample} & \multicolumn{4}{c|}{Test sample}\\
\cline{2-9}
\raisebox{1.5ex}[0cm][0cm]{Dataset}
& Vol. & Posit. & Negat. & Neutral & Vol. & Posit. & Negat. & Neutral\\
\hline
ROMIP-2013\footnotemark[3] & 4260 & 26 & 44 & 30 & 5500 & 32 & 41 & 27 \\
SRE-2015 Banks\footnotemark[4] & 6232 & 7 & 36 & 57 & 4612 & 8 & 14 & 78\\
SRE-2015 Telecom\footnotemark[4] & 5241 &  19 & 34 & 47 &  4173 & 10 & 23 & 67\\
SRE-2016 Banks\footnotemark[5] & 10725 & 7 &  26 & 67 &  3418 & 9 & 23 & 68\\
SRE-2016 Telecom\footnotemark[5] & 9209 & 15 &  28 & 57 & 2460 & 10 & 47 & 43\\
\hline
\end{tabular}
\end{table}

\section{Russian sentiment benchmark datasets}
In our study, we consider the following Russian datasets (benchmarks) annotated for previous Russian sentiment shared tasks: news quotes from the ROMIP-2013 evaluation \cite{chetviorkin2013evaluating} and  Twitter datasets from  SentiRuEval 2015-2016 evaluations \cite{loukachevitch2015entity,loukachevitch2016rubtsova}.  Table \ref{distributions} presents main characteristics of datasets including train and test sizes and  distributions by sentiment classes. It can be seen in Table \ref{distributions} that the neutral class is prevailing in all Twitter datasets. For this reason, along with the standard metrics of $F_1\ macro$ and accuracy, $F^{+-}_1macro$ and $F^{+-}_1micro$ ignoring the neutral class were also calculated. 

The collection of news quotes contains opinions in direct or indirect speech extracted from news articles \cite{chetviorkin2013evaluating}. The task of ROMIP-2013 evaluation was to distribute quotations between neutral, positive and negative classes depending on its sentiment. It can be seen in Table \ref{distributions} that dataset is rather balanced.

\footnotetext[3]{\url{http://romip.ru/en/collections/sentiment-news-collection-2012.html}}
\sloppy
\footnotetext[4]{\url{https://drive.google.com/drive/folders/1bAxIDjVz\_0UQn-iJwhnUwngjivS2kfM3}}
\footnotetext[5]{\url{https://drive.google.com/drive/folders/0BxlA8wH3PTUfV1F1UTBwVTJPd3c}}

Twitter datasets from SentiRuEval-2015-2016 evaluations were annotated for the task of reputation monitoring \cite{amigo2013overview,loukachevitch2015entity}, which means searching sentiment-oriented opinions about banks and telecom companies. In such a way this task can be regarded as an entity-oriented sentiment analysis problem. Insignificant part of samples contains two or more sentiment analysis objects, so these tweets are duplicated with corresponding attitude labels. The SentiRuEval-2016 training datasets are much larger in size as they contain training and test samples of 2015 evaluation \cite{loukachevitch2016rubtsova}. As it can be seen in Table \ref{distributions}, Twitter datasets are poorly balanced. This explains the choice of metrics considering only positive and negative classes.

\section{Automatic generation of annotated dataset}

The main idea of automatic annotation of dataset for targeted sentiment analysis task is based on the use of a sentiment lexicon comprising negative and positive words and phrases with their sentiment scores. We utilize Russian sentiment lexicon RuSentiLex \cite{loukachevitch2016creating}, which includes general sentiment  words of  Russian language, slang words from Twitter and words with positive or negative associations (connotations) from the news corpus. For ambigous words, having several senses with different sentiment orientations, RuSentiLex describes senses with references to the concepts of RuThes thesaurus \cite{loukachevitch2014ruthes}. The current version of RuSentiLex contains 16445 senses. 

As a  source for automatic dataset generation, we use a Russian news corpus, collected from various sources and representing different topics, which is important in fact that the benchmarks under analysis cover several topics. The corpus was collected long before the evaluations, so there are no possible overlaps between additional and benchmark data. The volume of the original corpus was about 4 Gb of raw text, which implies more than 10 million sentences.

The automatically annotated dataset includes general and thematic parts. For creation of the general part, we select monosemous positive and negative nouns from the RuSentiLex lexicon, which can be used as references to people or companies, which are sentiment targets in the benchmarks. We construct positive and negative word  lists and suppose that if a word from the list occurs in a sentence, it has a  context of the same sentiment. The list of positive and negative references to people or companies (seed words) includes 822 negative references and 108 positive ones. Examples  of such words are presented below (translated from Russian):
\begin{itemize}
    \item positive: \textit{"champion,  hero, good-looker"}, etc.;
    \item negative: \textit{"outsider,  swindler, liar, defrauder,  deserter"}, etc.
\end{itemize}

Sentences may contain several  seed words with different sentiments. In such cases, we duplicate sentences with labels in accordance with their attitudes.  The examples of extracted sentences are as follows (all further examples are translated from Russian): 
\begin{itemize}
\item positive: \textit{"A MASK is one who, on a gratuitous basis, helps the development of science and art, provides them with material assistance from their own funds"};
\item negative: \textit{"Such irresponsibility --- non-payments --- hits not only the MASK himself, but also throughout the house in which he lives"}.
\end{itemize}

To generate the thematic part of the automatic sample, we search for sentences that mention named entities depending on a task (banks or operators) using the named entity recognition model (NER) from DeepPavlov \cite{deeppavlov} co-occurred with sentiment words in the same sentences. We searched for sentences not only with organizations from benchmarks, but also with others companies from the relevant field. To ensure that a sentiment word refers to an entity, we restrict the distance between two words to be not more than four words. 

We remove examples containing a particle \textit{"not"} near sentiment word because it could change attitude of text in relation to target. Sentences with sentiment word located in quotation marks were also removed because they could distort the meaning of the sentence being a proper name.Examples of extracted thematic  sentiment sentences are as follows:

\begin{itemize}
    \item for banks (positive): \textit{"MASK increased its net profit in November by 10.7\%"}
    \item for mobile operators (negative): \textit{"FAS suspects MASK of imposing paid services on subscribers."}
\end{itemize}

Since the benchmarks contain also the neutral sentiment class, we need to extract sentences without sentiments. For this task, we choose among examples selected by NER those that do not contain any sentiment  words from the lexicon. Examples of extracted neutral sentences for both general and thematic parts are presented below:

\begin{itemize}
    \item for persons: \textit{"MASK is already starting training with its new team."}
    \item for banks: \textit{"On March 14, MASK announced that it was starting rebranding."}
    \item for mobile operators: \textit{"MASK has offered its subscribers a new service."}
\end{itemize}

 While creating an additional dataset, we take into account the distribution of sentiment words in the resulting sample, trying to bring it as close as possible to uniform. A source corpus contains enough examples with a negative sentiment to form a balanced dataset, which can not be said about words with the positive sentiment. We made automatically generated dataset publicly available\footnotemark[6].
 
 \footnotetext[6]{\url{https://github.com/antongolubev5/Auto-Dataset-For-Transfer-Learning}}

\begin{table}[h!]
\centering
\caption{Results based on training on additional dataset only.}\label{onlyadditionalequalbalanced}
\begin{tabular}{|l|l|c|c|c|c|}
\hline
Dataset & Model & Accuracy & $F_1\ macro$ & $F^{+-}_1macro$   & $F^{+-}_1micro$\\
\hline
& BERT-single & 28.32 & 21.54 & 45.74 & 46.19\\
ROMIP-2013 & BERT-pair-QA & 28.04 & 21.32 & 45.35 & 45.78\\
& BERT-pair-NLI & 27.76 & 20.89 & 45.12 & 45.68\\
\hline
& BERT-single & 33.42 & 25.10 & 39.17 & 42.29\\
SRE-2015 Banks & BERT-pair-QA & 33.19 & 25.56 & 38.98 & 42.31\\
& BERT-pair-NLI & 32.56 & 24.87 & 38.63 & 41.87\\
\hline
& BERT-single & 26.11 & 19.12 & 33.56 & 34.21\\
SRE-2015 Telecom & BERT-pair-QA & 26.12 & 19.05 & 32.61 & 34.43\\
& BERT-pair-NLI & 25.13 & 19.25 & 31.78 & 34.02\\
\hline
& BERT-single & 28.91 & 22.14 & 36.45 & 38.88\\
SRE-2016 Banks & BERT-pair-QA & 29.43 & 21.72 & 35.62 & 38.26\\
& BERT-pair-NLI & 28.58 & 20.42 & 34.38 & 37.73\\
\hline
& BERT-single & 25.86 & 19.57 & 32.87 & 34.59\\
SRE-2016 Telecom & BERT-pair-QA & 25.27 & 18.76 & 32.09 & 33.65\\
& BERT-pair-NLI & 24.14 & 18.23 & 31.06 & 33.28\\
\hline
\end{tabular}
\end{table}

\section{Text preprocessing}

To create an additional sample from the Russian news corpus, it was necessary to divide raw articles into separate sentences. For this task, we used rule-based sentence splitter from spaCy library \cite{spacy}, which is able to determine sentence boundaries automatically. This solution showed better quality in preliminary studies in comparison with NLTK variant \cite{nltk} and simple splitter based on regular expressions.

In addition to conceptual steps of creating an automatic dataset described in previous chapter, a few cleaning measures were performed. In accordance with calculated quantiles of sentences from test samples, too short and long examples were removed from additional data. To remove duplicate sentences from different sources, we use the metrics of cosine similarity between pairs of tf-idf representations of examples. When the value of the specified boundary value was exceeded, one of the sentences was randomly removed. Conducting experiments with different thresholds and exploring resulting samples, we set value equal to $0.8$.

After bringing the additional sample to the desired format, standard preprocessing track described in \cite{golubev2020loukachevitch},  including replacing similar text elements with appropriate tokens and removing special symbols  was carried out for all datasets.

\section{BERT architectures}

In our study, we consider three variants of fine-tuning BERT models \cite{devlin2018bert} for sentiment analysis. These architectures can be subdivided into the single-sentence approach using only initial text as an input and the two-sentence approach  \cite{utilizingbertforsa,golubev2020loukachevitch}, which converts the sentiment analysis task into a sentence-pair classification task by appending an additional sentence to the initial text.

The sentence-single model represents a vanilla BERT with an additional single linear layer on the top. The unique token \textit{[CLS]} is added for the classification task at the beginning of the sentence. The sentence-pair architecture adds an auxiliary sentence to the original input, inserting the \textit{[SEP]} token between two sentences. The difference between two models is in addition of a linear layer: for the sentence-pair model it is added over the final hidden state of \textit{[CLS]} token, while for the sentence-single variant it is added on the top of the entire last layer.

In our study, we use pre-trained Conversational RuBERT\footnotemark[7] from DeepPavlov framework \cite{rubert} trained on Russian social networks posts and comments which showed better results in preliminary study. 

\begin{table}[h!]
\centering
\caption{Results based on training on additional data mixed with benchmark train samples.}\label{additionalplustrain}
\begin{tabular}{|l|l|c|c|c|c|}
\hline
Dataset & Model & Accuracy & $F_1\ macro$ & $F^{+-}_1macro$   & $F^{+-}_1micro$\\
\hline
& BERT-single & 65.21 & 54.32 & 45.12 & 44.67\\
ROMIP-2013 & BERT-pair-QA & 65.53 & 54.68 & 45.73 & 45.14\\
& BERT-pair-NLI & 65.45 & 54.93 & 45.52 & 44.89\\
\hline
& BERT-single & 69.34 & 56.84 & 36.39 & 40.19\\
SRE-2015 Banks & BERT-pair-QA & 70.21 & 57.25 & 36.83 & 40.79\\
& BERT-pair-NLI & 69.54 & 57.06 & 36.65 & 40.31\\
\hline
& BERT-single & 66.43 & 53.19 & 33.41 & 37.71\\
SRE-2015 Telecom & BERT-pair-QA & 66.19 & 52.83 & 33.21 & 37.43\\
& BERT-pair-NLI & 67.11 & 53.48 & 33.73 & 38.03\\
\hline
& BERT-single & 67.71 & 54.76 & 33.61 & 37.85\\
SRE-2016 Banks & BERT-pair-QA & 67.61 & 54.85 & 34.53 & 36.89\\
& BERT-pair-NLI & 67.67 & 54.85 & 32.12 & 36.76\\
\hline
& BERT-single & 65.12 & 52.43 & 32.19 & 36.43\\
SRE-2016 Telecom & BERT-pair-QA & 64.76 & 52.06 & 32.28 & 36.12\\
& BERT-pair-NLI & 65.21 & 52.27 & 32.49 & 36.51\\
\hline
\end{tabular}
\end{table}

For the targeted sentiment analysis task, there are labels for each object of attitude so they can be replaced by a special token \textit{[MASK]}. Since general sentiment analysis problem has no certain 
attitude objects, token is assigned to the whole sentence and located at the beginning.

The sentence-pair model has two kind of architecture based on question answering (QA) and natural language inference (NLI) problems. The auxiliary sentences for each model are as follows:
\begin{itemize}
    \item pair-NLI: \textit{"The sentiment polarity of MASK is"}
    \item pair-QA: \textit{"What do you think about MASK?"}
\end{itemize}

\footnotetext[7]{\url{http://docs.deeppavlov.ai/en/master/features/models/bert.html}}

\section{Experiments and results}

We consider different options of constructing pre-training samples from the collected data and combining the resulting additional dataset with benchmark train samples. Different constructing variants comprise the following options:
\begin{itemize}
    \item training on the additional general and neutral thematic data  only and studying  dependence of the results on sentiment class distribution;
    \item training on the  additional general and neutral thematic data mixed with the benchmark training set;
    \item training on the full generated data (the data of previous steps are extended with sentiment-oriented  thematic examples) mixed with the benchmark training set;
    \item two-step approach: independent sequential training on additional dataset at the first step and on the benchmark training set at the second step;
    \item study of the dependence of the results on additional dataset size;
    \item three-step approach: independent sequential training in three stages using: the general data part from the additional dataset,  the thematic examples  from the additional dataset and the benchmark training sets.
\end{itemize}

All the results presented in the tables below are averaging over 3 experiments with different random initializations of models weights.

\subsection{Mixing additional data with  train samples}

As a starting point for research, we  train the models only on the automatically generated  dataset (general and thematic neutral sentences).  We compare two options of constructing the additional sample: uniform balancing between three sentiment classes and balancing in accordance with the average values of classes proportions for all datasets from Table \ref{distributions}. 

For both options, the sample size was chosen equal to 15000. The results obtained with uniform balancing are 2-3 \% higher and presented in Table \ref{onlyadditionalequalbalanced}. It can be seen, that performance is significantly lower than the current state-of-the-art results for all five benchmark datasets.

For the next step, we mix the automatically annotated data with the benchmark training sets. We keep the balance of sentiment classes from the previous experiment. The results are presented in Table  \ref{additionalplustrain}. For accuracy and $F_1\ macro$ metrics, the results improve significantly but still did not reach state-of-the-art level. It could be probably explained by assumptions about different topics and styles of texts in additional and benchmark datasets and time dependence of automatically generated data (too many sentences about sports and New Year celebration).

\begin{table}[h]
\centering
\caption{Results based on training on extended with sentiment thematic additional data mixed with the benchmark training sets.}\label{additionthematical}
\begin{tabular}{|l|l|c|c|c|c|}
\hline
Dataset & Model & Accuracy & $F_1\ macro$ & $F^{+-}_1macro$   & $F^{+-}_1micro$\\
\hline
& BERT-single & 66.78 & 62.44 & 71.49 & 70.61\\
ROMIP-2013 & BERT-pair-QA & 67.11 & 62.18 & 71.94 & 71.18\\
& BERT-pair-NLI & 67.89 & 63.24 & 72.27 & 71.65\\
\hline
& BERT-single & 70.54 & 66.18 & 67.31 & 66.59\\
SRE-2015 Banks & BERT-pair-QA & 70.87 & 66.71 & 68.24 & 66.91\\
& BERT-pair-NLI & 71.15 & 67.03 & 67.69 & 67.23\\
\hline
& BERT-single & 67.84 & 62.31 & 63.78 & 62.06\\
SRE-2015 Telecom & BERT-pair-QA & 68.35 & 62.44 & 64.21 & 62.51\\
& BERT-pair-NLI & 68.89 & 62.71 & 65.02 & 63.12\\
\hline
& BERT-single & 68.14 & 63.81 & 63.91 & 62.33\\
SRE-2016 Banks & BERT-pair-QA & 68.81 & 64.42 & 65.43 & 64.16\\
& BERT-pair-NLI & 69.21 & 65.02 & 65.76 & 65.59\\
\hline
& BERT-single & 67.31 & 62.15 & 63.28 & 61.68\\
SRE-2016 Telecom & BERT-pair-QA & 67.59 & 62.31 & 63.46 & 62.01\\
& BERT-pair-NLI & 68.16 & 63.37 & 64.19 & 62.21\\
\hline
\end{tabular}
\end{table}
 
\subsection{Extension of additional sample by thematic data}
Analyzing low results of the previous experiment, we supposed it may be associated with topic differences  between automatic and benchmark datasets, since at this stage an automatic sample was collected using personal descriptive words only. This way, we extend the additional dataset with sentiment thematic examples using the list of well-known organizations (banks and operators) and sentences obtained with NER from DeepPavlov, keeping sample size and sentiment class ratio unchanged. 

The results are presented in Table \ref{additionthematical}. For all $F_1$ metrics, the performance seems much better than in the previous experiment (mixed general additional sample and training benchmark datasets), but still worse than current state-of-the-art results.

\subsection{Two-step transfer learning approach}

The two-step transfer learning  consists in the sequential training on two samples and differs from the previous one in that we do not mix automatically generated data with benchmarks train sets. At the first step, the models are trained on the additional data, then model weights are frozen and training continues on the training data from the benchmarks.

During the same experiment, we  study the dependence between the results and size of additional dataset. It was found that with increasing sample size, the results improve too. The boundary between extension of additional dataset and increasing the results was set at a sample size of 27000 (9000 per each sentiment class). Using the two-step approach allows us to overcome the current best results for almost all datasets. The results of described experiment and comparison with the state-of-the-art results \cite{smetanin2021deep,golubev2020loukachevitch} are presented in Table \ref{2step}. 

\begin{table}[h!]
\centering
\caption{Results based on using the two-step approach.}\label{2step}
\begin{tabular}{|l|l|c|c|c|c|}
\hline
Dataset & Model & Accuracy & $F_1\ macro$ & $F^{+-}_1macro$   & $F^{+-}_1micro$\\
\hline
& BERT-single & 79.95 &  71.16 & 85.39 & 85.61\\
ROMIP-2013 & BERT-pair-QA & 80.21 &  71.29 & 85.72 & 85.93\\
& BERT-pair-NLI & \textbf{80.56} &  \textbf{71.68} & \textbf{86.14} & \textbf{86.19}\\
& Current SOTA & 80.28 & 70.62 &  85.52 &  85.68\\
\hline
& BERT-single & 86.06 & 79.11 & 64.87 & 66.73\\
SRE-2015 Banks & BERT-pair-QA & 86.34 & 79.58 & 65.29 & 67.02\\
& BERT-pair-NLI & \textbf{87.62} & \textbf{80.72} & \textbf{68.44} & \textbf{71.39}\\
& Current SOTA & 86.88 & 79.51 & 67.44 & 70.09\\
\hline
& BERT-single & 77.11 & 69.76 & 61.89 & 66.95\\
SRE-2015 Telecom & BERT-pair-QA & \textbf{78.14} & \textbf{70.03} & \textbf{64.53} & \textbf{68.29}\\
& BERT-pair-NLI & 77.96 & 69.68 & 64.52 & 68.21\\
& Current SOTA & 76.63 & 68.54 & 63.47 & 67.51\\
\hline
& BERT-single & 81.94 & 74.08 & 67.24 & 70.68\\
SRE-2016 Banks & BERT-pair-QA & \textbf{84.36} & \textbf{77.43} & \textbf{72.32} & \textbf{74.06}\\
& BERT-pair-NLI &  84.19 & 75.63 & 68.52 & 70.89\\
& Current SOTA & 82.28 & 74.06 & 69.53 & 71.76\\
\hline
& BERT-single & 75.82 & 69.78 & 65.04 & 74.22\\
SRE-2016 Telecom & BERT-pair-QA & 77.25 & 69.71 & 67.35 & 76.22\\
& BERT-pair-NLI &  \textbf{77.59} & 69.84 & \textbf{68.11} & 75.93\\
& Current SOTA & -- & \textbf{70.68} & 66.40 & \textbf{76.71}\\
\hline
\end{tabular}
\end{table}

\subsection{Three-step transfer learning approach}

For the final experiment of the study, we  divide the first step of the previous experiment into two: sequential training on the general and thematic data. At first, the  models are trained on the general data, then the weights are frozen and the training continues on the thematic examples retrieved with the list of organizations and NER from DeepPavlov. After the second weights freezing, the last stage of learning on the original training samples begins. Taken together, this sequence represents the three-step transfer learning approach.

During this experiment, we also changed the additional sample by adding sentiment examples to thematic part of additional sample. The logic consisted in the selection among thematic sentences, those which contain sentiment words. Thus, the first step sample contains 18000 general examples and the second sample consists of 9000 thematic examples (both samples are equally balanced across sentiment classes).

The use of three-step approach combined with addition of sentiment thematic contexts to the sample, improved the results by a few more points. New state-of-the-art results as well as comparison with manual labelling for SentiRuEval-2015 telecom dataset are presented in Table \ref{3step}. According to the organizers of SentiRuEval-2016 evaluation, one participant sent the results of manual annotation of the test set \cite{loukachevitch2016rubtsova}. As it can be seen, BERT-pair-NLI model reaches human sentiment analysis level by $F^{+-}_1micro$ metric.

\begin{table}[h]
\centering
\caption{Results based on the  three-step approach.}\label{3step}
\begin{tabular}{|l|l|c|c|c|c|}
\hline
Dataset & Model & Accuracy & $F_1\ macro$ & $F^{+-}_1macro$   & $F^{+-}_1micro$\\
\hline
& BERT-single & 80.27 &  71.78 & 85.82 & 86.07\\
ROMIP-2013 & BERT-pair-QA & 80.78 &  72.09 & 86.14 & 86.42\\
& BERT-pair-NLI & \textbf{82.33} &  \textbf{72.69} & \textbf{86.77} & \textbf{87.04}\\
& Current SOTA & 80.28 & 70.62 &  85.52 &  85.68\\
\hline
& BERT-single & 87.65 & 80.79 & 65.74 & 67.46\\
SRE-2015 Banks & BERT-pair-QA & 87.92 & 81.12 & 66.47 & 68.55\\
& BERT-pair-NLI & \textbf{88.14} & \textbf{81.63} & \textbf{68.76} & \textbf{72.28}\\
& Current SOTA & 86.88 & 79.51 & 67.44 & 70.09\\
\hline
& BERT-single & 77.85 & 70.42 & 62.29 & 67.38\\
SRE-2015 Telecom & BERT-pair-QA & \textbf{79.21} & 70.94 & 65.68 & 69.11\\
& BERT-pair-NLI & 79.12 & \textbf{71.16} & \textbf{65.71} & \textbf{70.65}\\
& Current SOTA & 76.63 & 68.54 & 63.47 & 67.51\\
& Manual \cite{loukachevitch2016rubtsova} & -- & -- & 70.30 & 70.90\\
\hline
& BERT-single & 83.21 & 75.31 & 68.45 & 71.69\\
SRE-2016 Banks & BERT-pair-QA & \textbf{85.59} & \textbf{78.93} & \textbf{74.05}  & \textbf{75.12}\\
& BERT-pair-NLI &  85.43 & 76.85 & 70.23 & 72.07\\
& Current SOTA & 82.28 & 74.06 & 69.53 & 71.76\\
\hline
& BERT-single & 76.79 & 70.64 & 66.16 & 75.27\\
SRE-2016 Telecom & BERT-pair-QA & 78.42 & 70.54 & \textbf{68.65} & \textbf{77.45}\\
& BERT-pair-NLI &  \textbf{78.62} & \textbf{71.18} & 69.36 & 76.85\\
& Current SOTA & -- & 70.68 & 66.40 & 76.71\\
\hline
\end{tabular}
\end{table} 

\section{Conclusion}

In this study, we presented a method for automatic generation of annotated sample from a Russian news corpus using distant supervision technique. We compared different options of combining additional data with benchmark train samples and improved current state-of-the-art results by  more  than  3\% using BERT models together with the transfer learning approach. The best variant was three-step approach of sequential training on general, thematic and benchmark train samples with intermediate freezing of the model weights. On one of benchmarks, the BERT-NLI model treating a sentiment classification problem as a natural language inference task, reached human level according to one of the metrics.

\section*{Acknowledgments}
The reported study was funded by RFBR according to the research project \textnumero~20-07-01059.

\bibliographystyle{ugost2008ls}

\end{document}